\documentclass[letterpaper, 10 pt, conference]{ieeeconf}  

\IEEEoverridecommandlockouts                              

\usepackage{amsmath,amssymb,amsfonts}
\usepackage{algorithmic}
\usepackage{graphicx} 
\usepackage[dvipsnames]{xcolor}
\usepackage{textcomp}
\usepackage{xcolor}
\usepackage[baseline]{euflag}
\usepackage{hyperref}
\usepackage{siunitx}
\usepackage{orcidlink}

\usepackage{adjustbox}
\usepackage[ruled,vlined]{algorithm2e}
\usepackage[font=footnotesize]{caption}
\usepackage{subcaption}
\usepackage{float}
\usepackage{multirow}
\usepackage{fancyhdr}

\title{\LARGE \bf
Self-Supervised Data Generation for Precision Agriculture: Blending Simulated Environments with Real Imagery
}

\author{Leonardo Saraceni\orcidlink{0000-0001-5817-3893}$^{1,*}$, Ionut M. Motoi\orcidlink{0000-0002-9562-6799}$^{1}$, Daniele Nardi\orcidlink{0000-0001-6606-200X}$^{1}$ and Thomas A. Ciarfuglia\orcidlink{0000-0001-8646-8197}$^{1}$
\thanks{\euflag \quad This work is part of a project that has received funding from the European Union’s Horizon 2020 research and innovation programme under grant agreement No 101016906 – Project CANOPIES}
\thanks{\euflag \quad This work has been partially supported by project AGRITECH Spoke 9
- Codice progetto MUR: AGRITECH ”National Research Centre for Agricultural
Technologies” - CUP CN00000022, of the National Recovery and Resilience Plan
(PNRR) financed by the European Union ”Next Generation EU”.}%
\thanks{This work has been partially supported by Sapienza University of Rome as part of the work for project \textit{H\&M: Hyperspectral and Multispectral Fruit Sugar Content Estimation for Robot Harvesting Operations in Difficult Environments}, Del. SA n.36/2022.}%
\thanks{$^{1}$All the authors are with the Department of Computer, Control and Management Engineering (DIAG), Sapienza University of Rome, via Ariosto 25, Rome, Italy.}
\thanks{$^{*}$ Corresponding author saraceni@diag.uniroma1.it}%
\thanks{}
\thanks{979-8-3503-5663-2/24/\$31.00~\copyright2024 European Union}%
}

\begin{document}

\maketitle

\begin{abstract}
In precision agriculture, the scarcity of labeled data and significant covariate shifts pose unique challenges for training machine learning models. This scarcity is particularly problematic due to the dynamic nature of the environment and the evolving appearance of agricultural subjects as living things. We propose a novel system for generating realistic synthetic data to address these challenges. Utilizing a vineyard simulator based on the Unity engine, our system employs a cut-and-paste technique with geometrical consistency considerations to produce accurate photo-realistic images and labels from synthetic environments to train detection algorithms. This approach generates diverse data samples across various viewpoints and lighting conditions. We demonstrate considerable performance improvements in training a state-of-the-art detector by applying our method to table grapes cultivation. The combination of techniques can be easily automated, an increasingly important consideration for adoption in agricultural practice.
\end{abstract}


\section{Introduction}

The application of robotics in precision agriculture is rapidly advancing, alongside the use of data-driven algorithms for various tasks \cite{cheng2023recent,droukas2023survey}. In particular, Computer Vision (CV) has become integral for many monitoring tasks, such as detection for yield estimation \cite{kurtser2020in-field}, pest and illnesses control \cite{lacotte2022pesticide-free}, and visual servoing for harvesting or spraying \cite{pretto2020building, lippi2023autonomous, stavridis2121pick-and-place}. However, the agricultural domain presents unique challenges due to its constantly changing environment caused by normal agricultural operations and the organic nature of the observed subjects. 

\begin{figure}[t]
\centering
    \begin{subfigure}[]{0.475\textwidth}
        \includegraphics[width=\columnwidth]{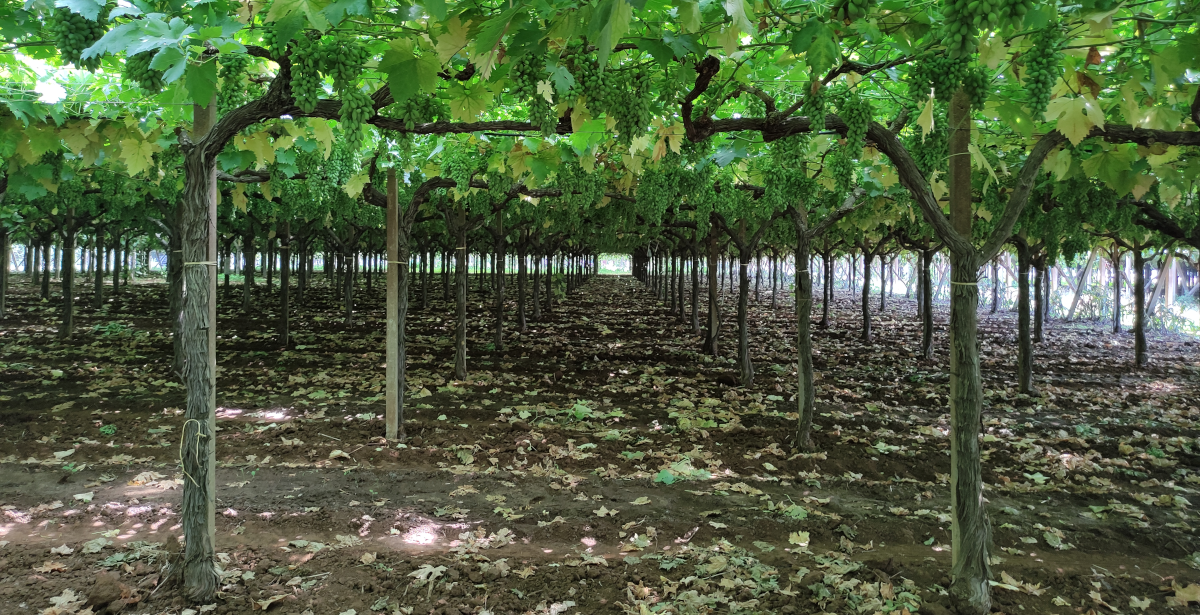}
        \caption{}
        \label{fig:tendone}
    \end{subfigure}
    \begin{subfigure}[]{0.185\textwidth}
        \includegraphics[width=\columnwidth]{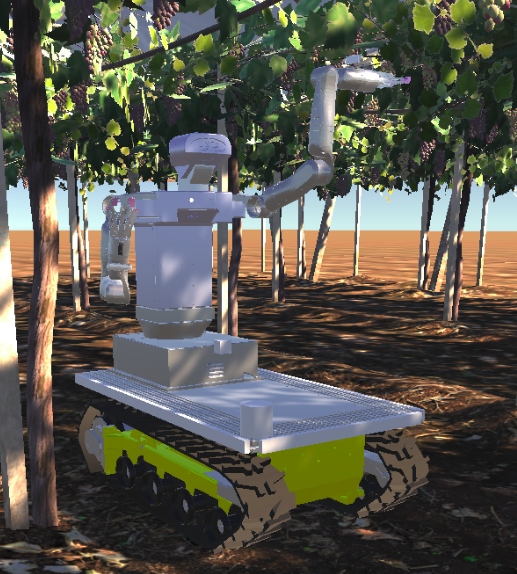}
        \caption{}
        \label{fig::robot}
    \end{subfigure}
    \begin{subfigure}[]{0.29\textwidth}
        \includegraphics[width=\columnwidth]{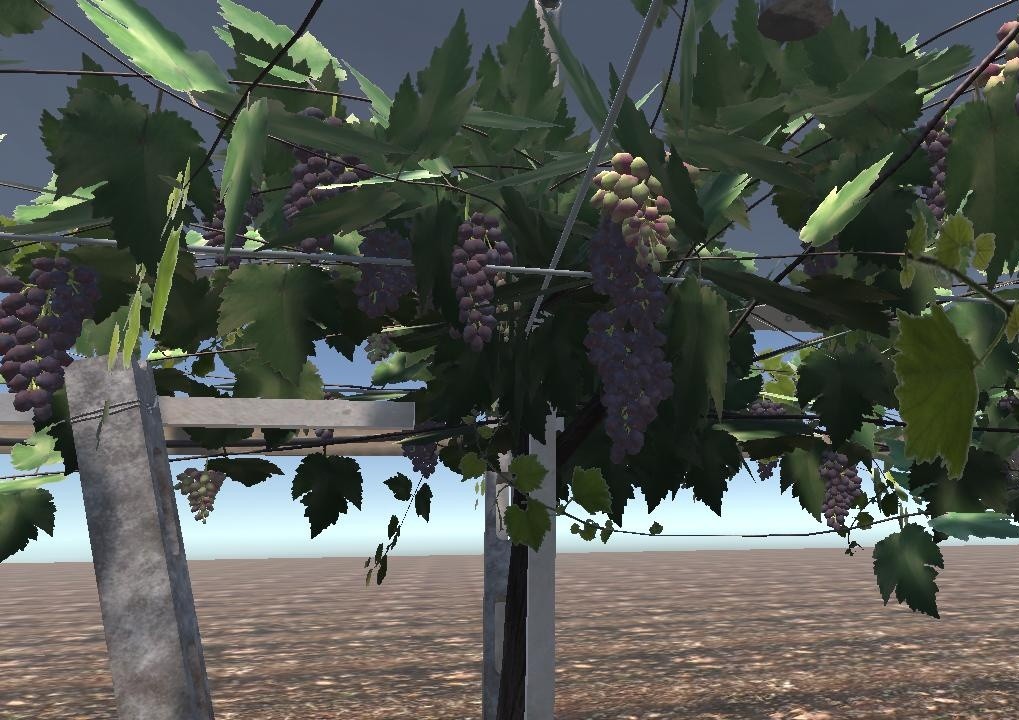}
        \caption{}
        \label{fig:simulated-grapes}
    \end{subfigure}
    \caption{\footnotesize a) The adopted operational environment in Aprilia (Lazio). b) Robotic platform used in the EU Project CANOPIES placed in the simulated environment. c) Synthetic image of the grapes captured from the robot camera point of view.}
    \label{fig::simulator}
\end{figure}

\begin{figure*}[t]
\footnotesize
\centering
\includegraphics[width=\textwidth]{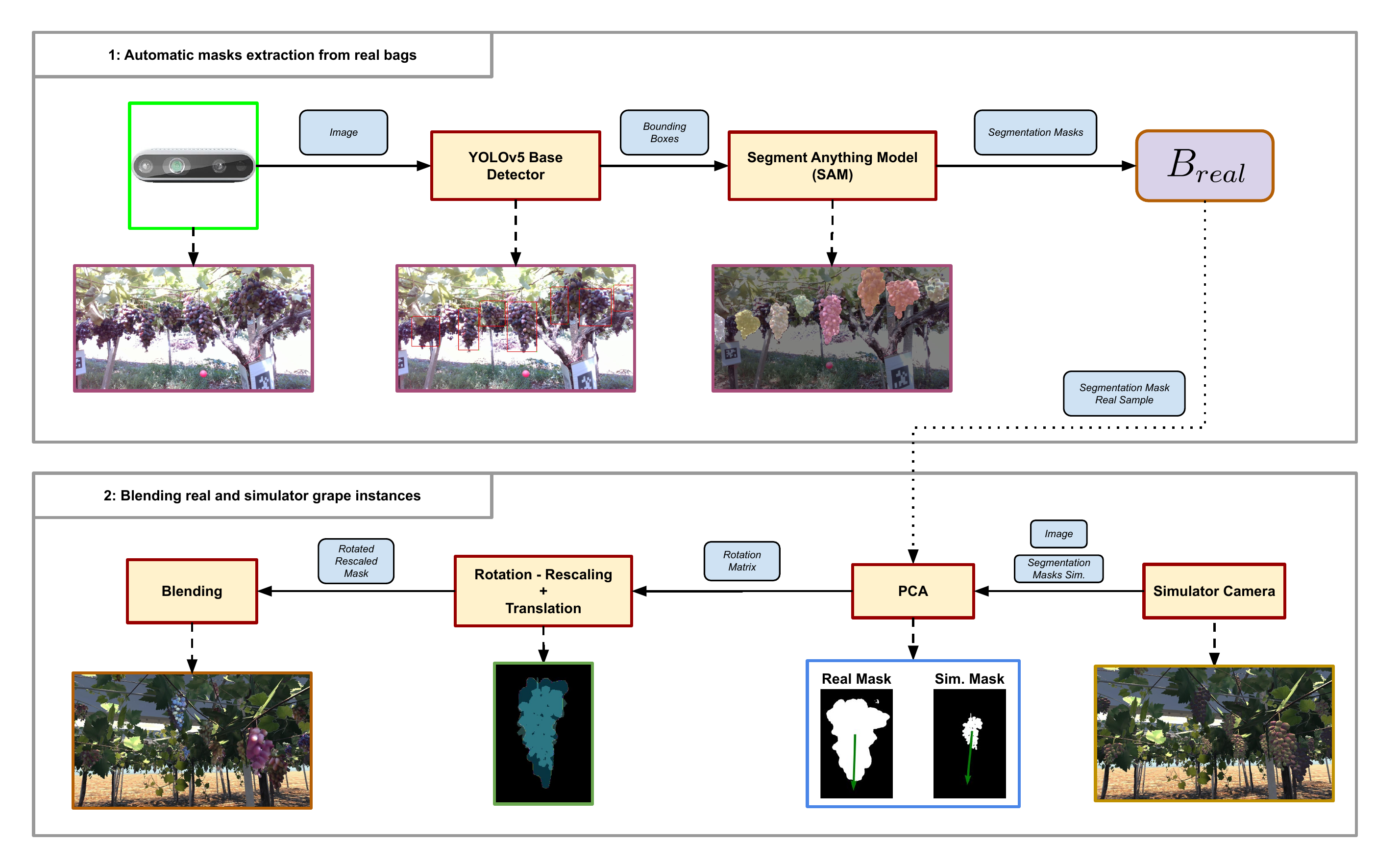}

\caption{\footnotesize The proposed pipeline is divided into two parts. 1) The base detector (Yolov5) extracts bounding boxes from real images captured in the vineyard. We use those as input prompts for SAM to extract the segmentation masks of single instances, which we save in a buffer $B_{real}$. 2) For every grape instance in the synthetic images, we randomly sample a real mask from $B_{real}$ and perform PCA to align them. Then, we rescale and translate the real instance to overlap and blend it with the synthetic one.}
\label{fig::pipeline}
\end{figure*}


Since the development and implementation of robust detection and segmentation models depend heavily on the availability of high-quality, annotated datasets, agricultural settings pose specific challenges. Often, there is no way to control illumination in the field since the sun's position depends on the time of the day and the weather conditions. Foliage and other obstacles can cause sudden illumination shifts, with the consequence of over-exposing the camera sensor. Motion blur is very common due to the uneven terrain over which the robot is moving. In addition, fruits such as table grapes are also self-similar and self-occluding since they grow very close to each other. All these issues make the detection task more complex than in other environments. Finally, since the target is a living organism, the phenomena of covariate shifts, where the statistical properties of training data differ from those encountered in operational environments, is another considerable challenge to the generalization of data-driven algorithms. While approaches to detect an out-of-distribution decrease in performance exist \cite{ginart2022mldemon, park2021reliable}, the problem of collecting new labeled data to retrain the models remains open.

Many different approaches exist for data generation and augmentation in CV. However, for adoption in agriculture, they should be simple enough to be operated by farmers or agronomists, who generally lack specialized computer science skills. For this reason, agronomic data generation approaches to compensate for covariate shifts should be automated, allowing users to interact with the technology primarily on an agronomic level.

This work follows this line by proposing a system that automatically generates data for detection and tracking. These are two typical CV tasks on which other more complex agronomic tasks rely, like harvesting, yield estimation, and pest detection. The only requirement of our method is that the farmer or agronomist selects some initial samples from images collected in the field that are representative of the actual crop of the season. We use the case of table grapes as an example of a crop that is subjected to covariate shifts in appearance from one year to the following and whose training system, where the vine leaves create a roof for the fruits (Fig. \ref{fig:tendone}), proves to be a very challenging one due to uneven illumination.

The proposed system uses a 3D vineyard simulator that was developed in the context of the EU project CANOPIES \cite{canopies} (Fig. \ref{fig::robot} and \ref{fig:simulated-grapes}). This virtual environment was built to provide a testing ground for robotic navigation and human-robot interaction but does not have a photo-realistic appearance, and it can be considered a typical 3D environment that can be built by using commonly available graphics assets without requiring refined 3D modeling skills. The simulator allows for extracting camera sequences and labeling fruits with bounding boxes and segmentation masks. The extracted fruits can be used to enrich an initial dataset for detecting fruits, but, given their simple appearance, it does not help in facing the covariate shift problem. For this reason, we propose to use real images collected on the field and a combination of weak detectors and semi-supervised zero-shot learning to segment real grape images and then paste them with different strategies onto the synthetic ones to create new samples that overcame the illumination and seasonal shifts. The work's main contribution is the proposed pipeline, which can be replicated for other crops. In addition, we release the data and the software used in the specific case of table grapes.

The following Section \ref{sec::related} gives a brief overview of the synthetic data generation efforts in agriculture, while the proposed method and the data used are described in Section \ref{sec::methods}. Section \ref{sec::experiments} describes the comparative experiments with a state-of-the-art detector trained using standard data and augmented ones. Finally, Section \ref{sec::conclusions} draws the conclusions.

\begin{figure*}[t]
\centering
	\begin{subfigure}[]{0.49\textwidth}
		\includegraphics[width=\columnwidth]{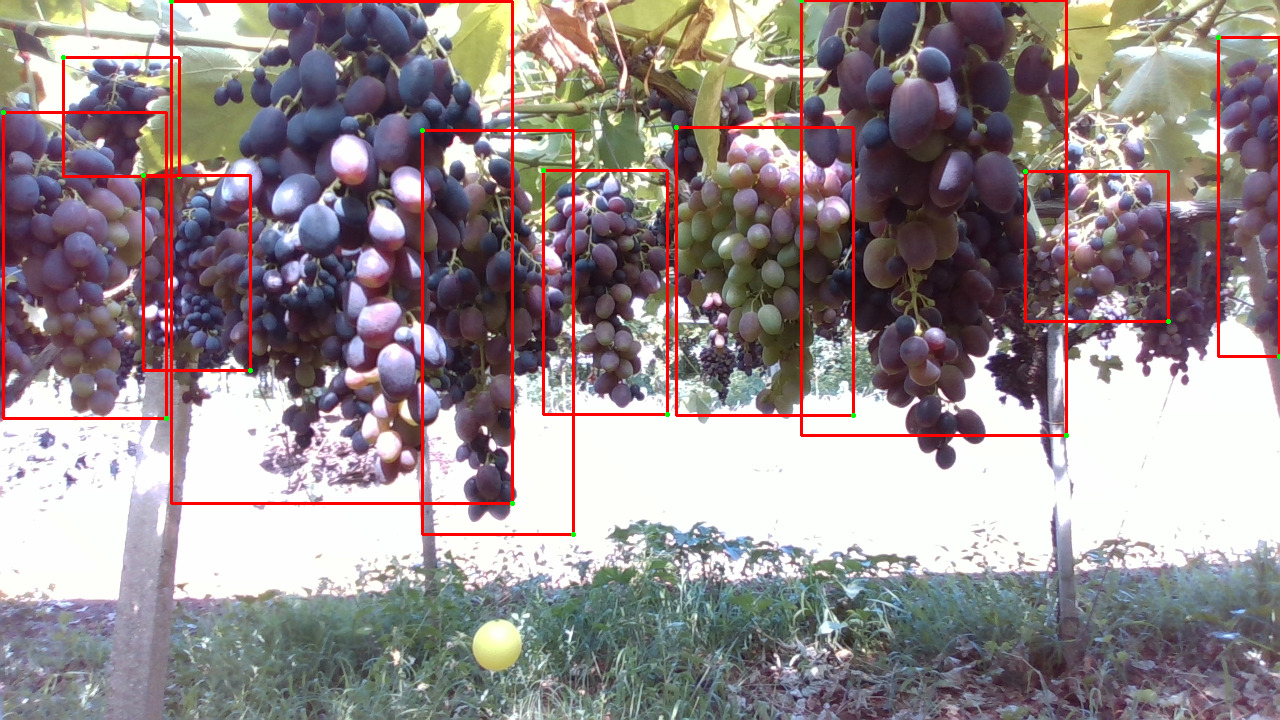}
		\caption{}
	\end{subfigure}
        \begin{subfigure}[]{0.49\textwidth}
		\includegraphics[width=\columnwidth]{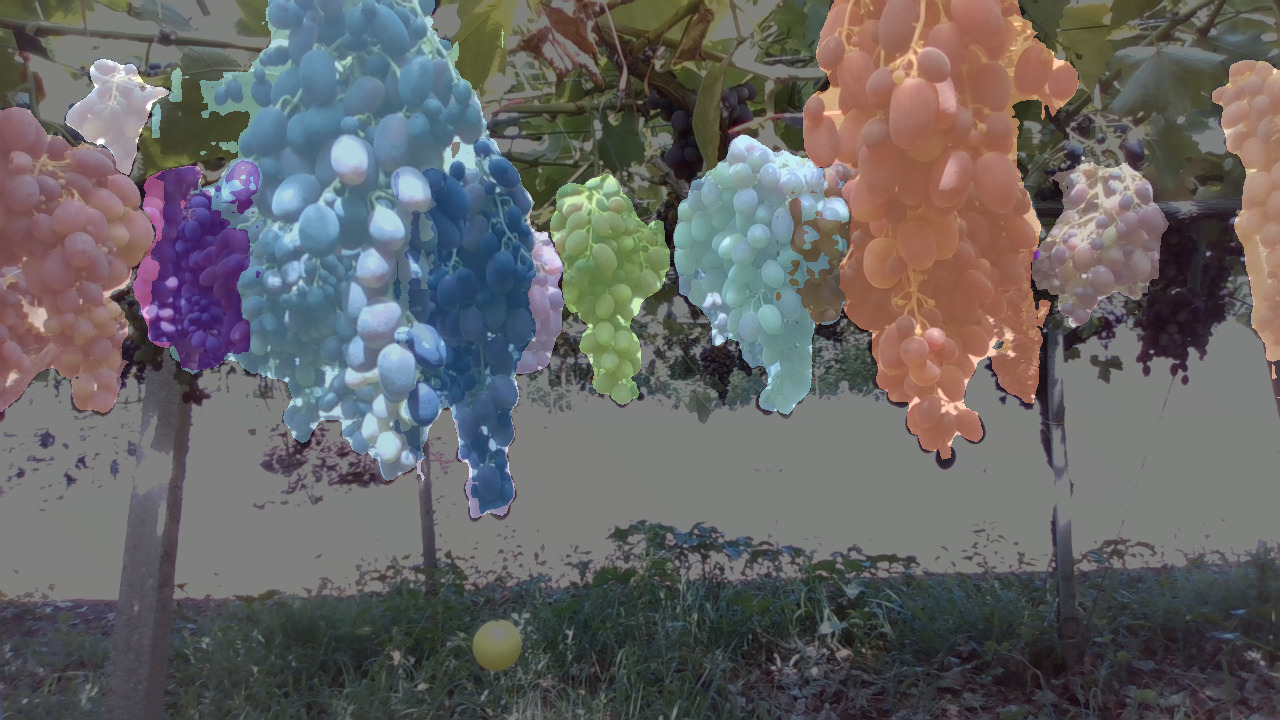}
		\caption{}
	\end{subfigure}
	\caption{\footnotesize Example of automatic detection using the base detector trained using automatically generated pseudo-labels (a) and segmentation by SAM using the detection as input prompt (b).}
\label{fig::segmentation}
\end{figure*}

\section{Related Work} 
\label{sec::related}
A broad analysis of the implications of applying robotics and data-driven technologies to agriculture can be found in \cite{roscher2023data}. The authors stress how, in more than four decades of Robotics advancements in solving agricultural problems, a broad adoption in the agricultural practice is still lacking \cite{kamilaris2017review}. In the study, different reasons are brought up. However, one of the most relevant is the unbalanced cost-benefit trade-off in current practices for data collection, creation, curation, and use that is critical for the generalization power of data-driven algorithms. 

It is not surprising that, among the wide spectrum of research in digital agriculture, data generation with self and semi-supervised techniques has attracted much attention. In a recent review on the topic \cite{li2023label}, the authors count more than 50 articles since 2016 on label-efficient learning. Among all these approaches, generative sampling approaches are the ones that try to generate new labeled samples starting from a subset of the existing ones or by generating them from scratch. In this respect, a technology that is frequently exploited for realistic artificial sample generation is Generative Adversarial Networks (GANs) \cite{goodfellow2020generative}. A detailed review can be found in \cite{lu2022generative}. These models have been successfully used to generate samples in a range of very different agronomic scenarios, such as plant seedlings images \cite{madsen2019generating}, \textit{arabidopsis} images \cite{valerio2017arigan}, soil moisture images \cite{hammouch2022ganset} and whiteflies pest images \cite{karam2022gan-based} to mention a few examples. However, these networks are notoriously difficult to train and require curated training datasets and specialized experience \cite{lu2022generative}. 

On the other hand, direct data synthesis with graphics engines and CAD software is another common strategy to address data scarcity in Computer Vision tasks \cite{paulin2023review}. In agriculture, some authors tried to leverage modern simulation engines to extract relevant data and automatize labeling \cite{barth2018data}. However, while many free resources exist for modeling simple natural scenes, creating more realistic environments is expensive and requires highly skilled work. 

Given the limitations of 3D engines in terms of realism, some authors put together both CAD-generated samples and GANs \cite{hartley2021domain, barth2018improved} using CycleGANs \cite{zhu2017unpaired}. 
In all cases, the combined strategy can provide a considerable performance increase, but again, it is difficult to automate due to the training instabilities of GANs.

Compared to the works mentioned above, we also use a synthetic environment to generate new samples. However, we avoid needing expert data scientists or skilled graphics engineers to train complex generative models. Instead, we use simple, easily automatized CV algorithms to provide a complete generation pipeline with a degree of realism sufficient to give consistent performance improvements.

\section{Materials and Methods}
\label{sec::methods}

This section describes the pipeline for the self-supervised generation of realistic data from synthetic sources. Section \ref{sec::data} provides the process for acquiring the synthetic data and a short description of the simulator. In sections \ref{sec::segmentation} and \ref{sec::paste}, we discuss the automatic extraction of real-world segmentation data and the method for the pasting process in the simulated environment. The pipeline of the process is presented in Fig. \ref{fig::pipeline}.

\subsection{Synthetic Data Acquisition}
\label{sec::data}
The Unity-based simulator \cite{canopiesD2.4} developed for the CANOPIES Project reproduces a traditional vineyard trellis system called Tendone, with a distance between each plant of 3 meters. The robot (Fig. \ref{fig::robot}) represented in the simulation environment is a tracked Alitrak DCT-300P mobile base mounting a modified PAL Robotics Thiago++ humanoid manipulator on top. The end-effector of the robotic platform is equipped with a RealSense d435i depth camera, which provides both the color image and the ground truth segmentation image with HD resolution (1280x720). 
The simulator allows for the configuration of the number, size, and density of table grape bunches in the scene. In addition, the position of the main light source (the Sun) can be set. The simulator is ROS-compatible, which makes data acquisition simple. 
We move the robot along the vineyard rows with the arm-mounted camera directed towards the vine trees, simulating a typical viewpoint during harvesting and monitoring operations (Fig. \ref{fig:simulated-grapes}). We ensure maximum variability and robustness to the phenomena of over-exposure of the camera sensor by iteratively tuning the simulator's lighting parameter using four equally spaced numbers in the range between the minimum and maximum allowable values.

\begin{figure}[t]
\centering
    \includegraphics[width=\columnwidth]{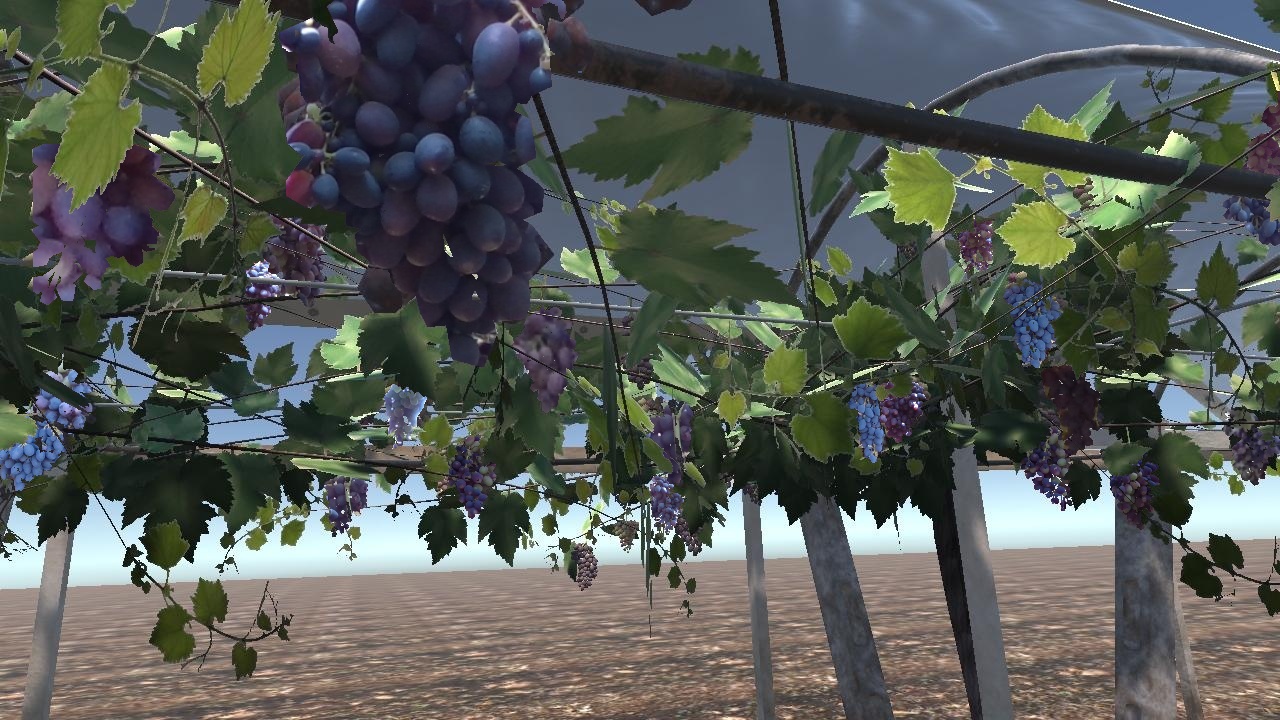}
    \caption{\footnotesize Example of simulator image blended with real grape instances using the pasting method described in Section \ref{sec::paste}}
    \label{fig::setup}
\end{figure}

\subsection{Real World Self-Supervised Segmentation}
\label{sec::segmentation}
The main strategy of this work is to take segmentation masks from real images and fuse them in the simulator to provide more variability in the data distribution and train more generalized detection models. To avoid the cost of manual segmentation and to provide a fully automatic solution, we exploit a base table grape detector based on YOLOv5 trained using automatically generated pseudo-labels \cite{ciarfuglia2023weakly, ciarfuglia2022pseudo}, and Segment Anything Model (SAM) \cite{kirillov2023segment}, a foundational visual model that allows for zero-shot segmentation using both visual and textual prompts.
In our case, we use the bounding boxes extracted by the base detector as the input prompt to SAM, which automatically extracts segmentation masks (\textit{e.g.} Fig. \ref{fig::segmentation}). We save the labels corresponding to the extracted segmentation instances in a buffer $B_{real}$ to prepare them for the next step: pasting them in the virtual environment images on top of the grape instances on the simulator images.

\begin{figure*}[t]
\centering
    \begin{subfigure}[]{0.325\textwidth}
        \centering
        \caption*{\normalsize \textbf{Pseudo}}
        \includegraphics[width=\columnwidth]{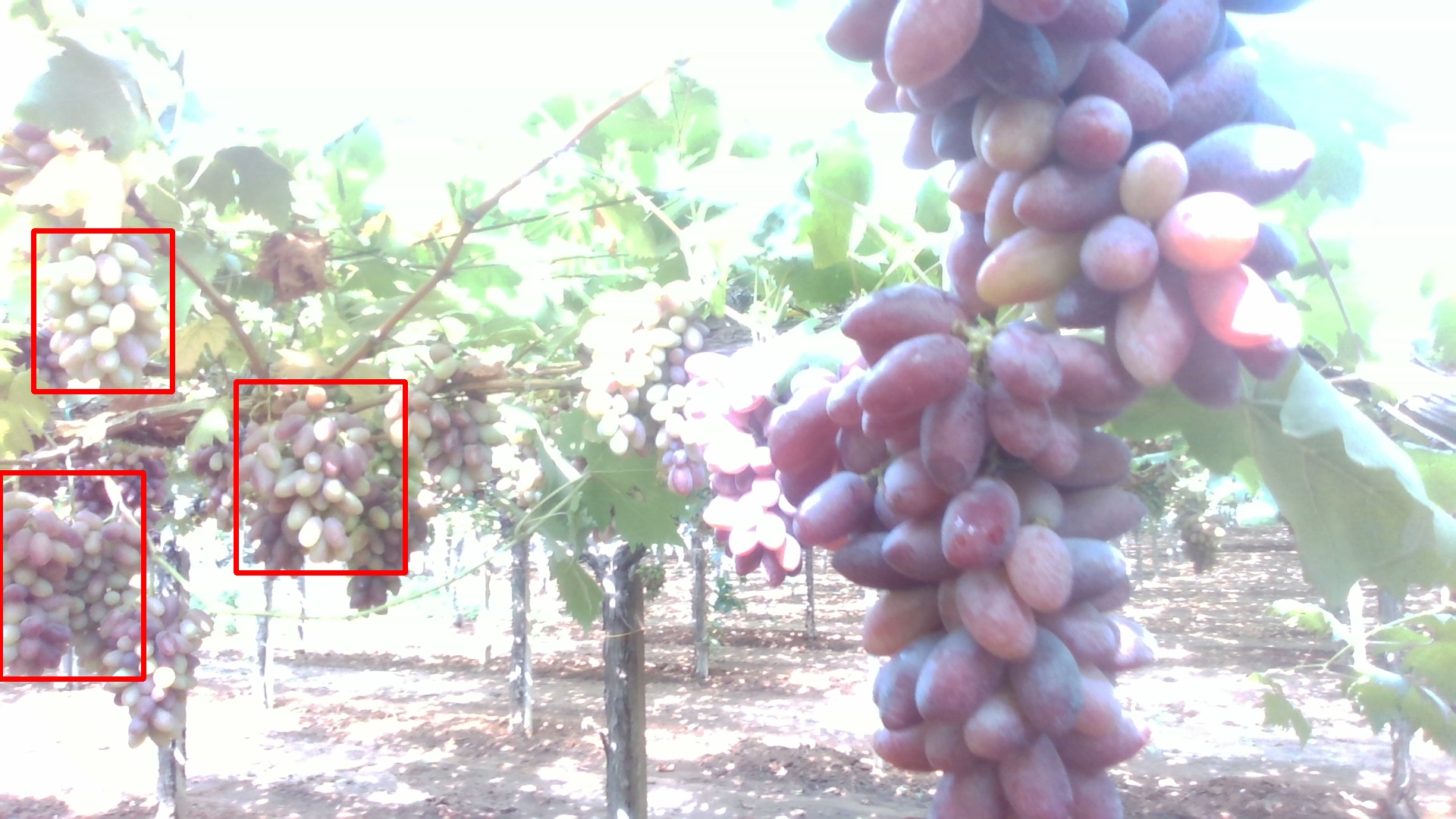}
    \end{subfigure}
    \begin{subfigure}[]{0.325\textwidth}
        \centering
        \caption*{\normalsize \textbf{Synthetic + Pseudo}}
        \includegraphics[width=\columnwidth]{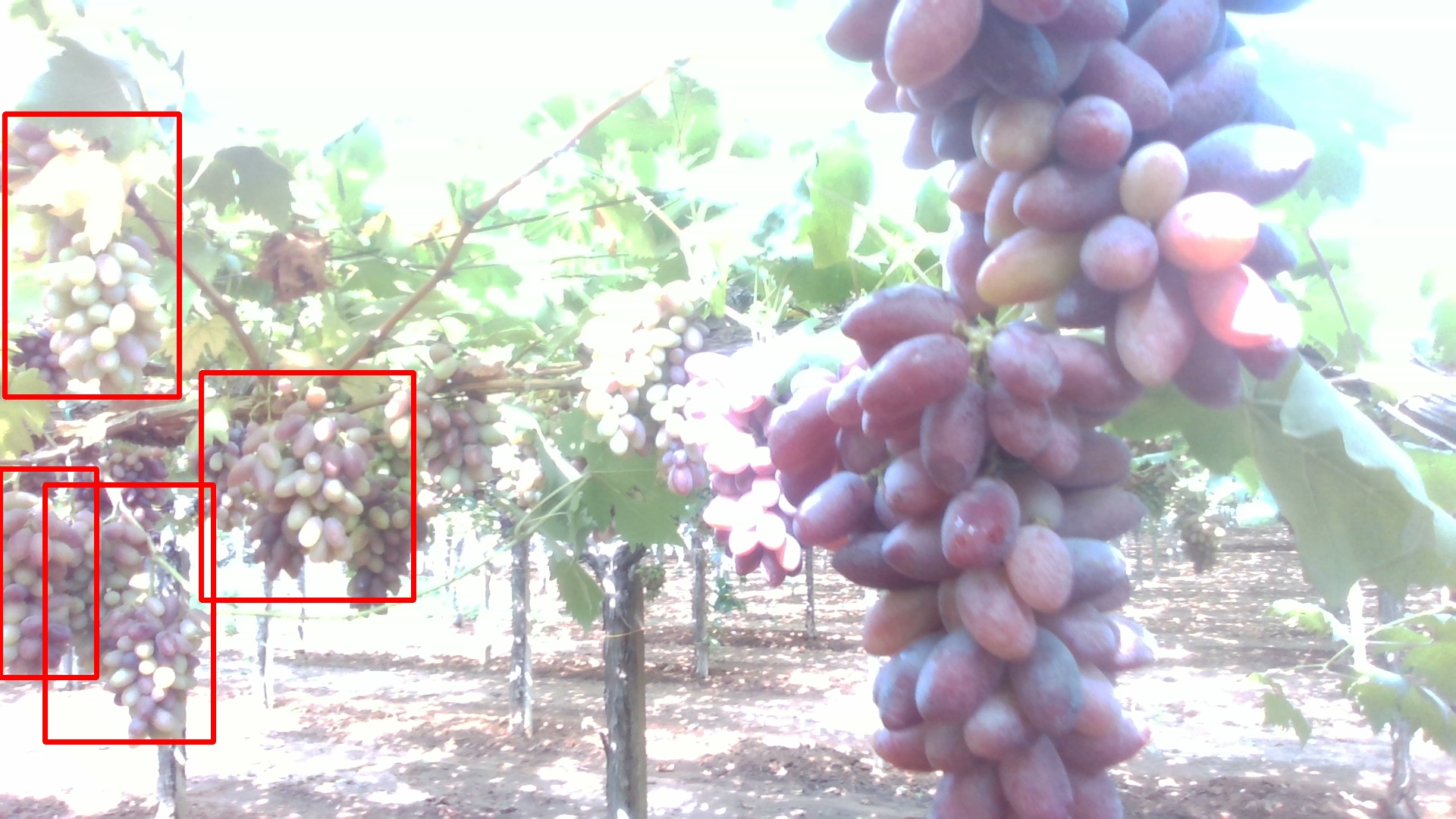}
    \end{subfigure}
    \begin{subfigure}[]{0.325\textwidth}
        \centering
        \caption*{\normalsize \textbf{Synthetic Pasted + Pseudo}}
        \includegraphics[width=\columnwidth]{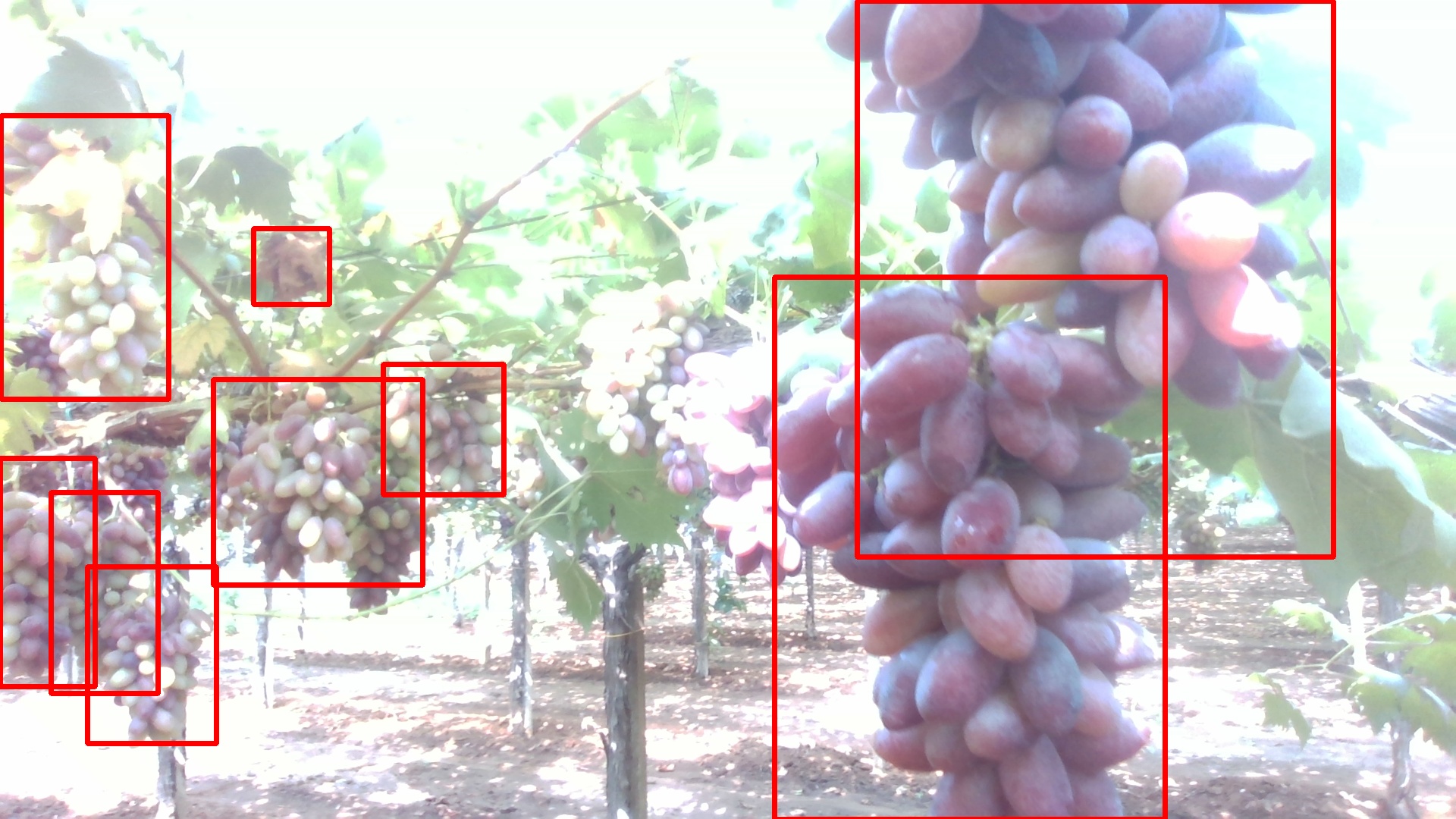}
    \end{subfigure}

    \vspace{5pt}
    \begin{subfigure}[]{0.325\textwidth}
        \includegraphics[width=\columnwidth]{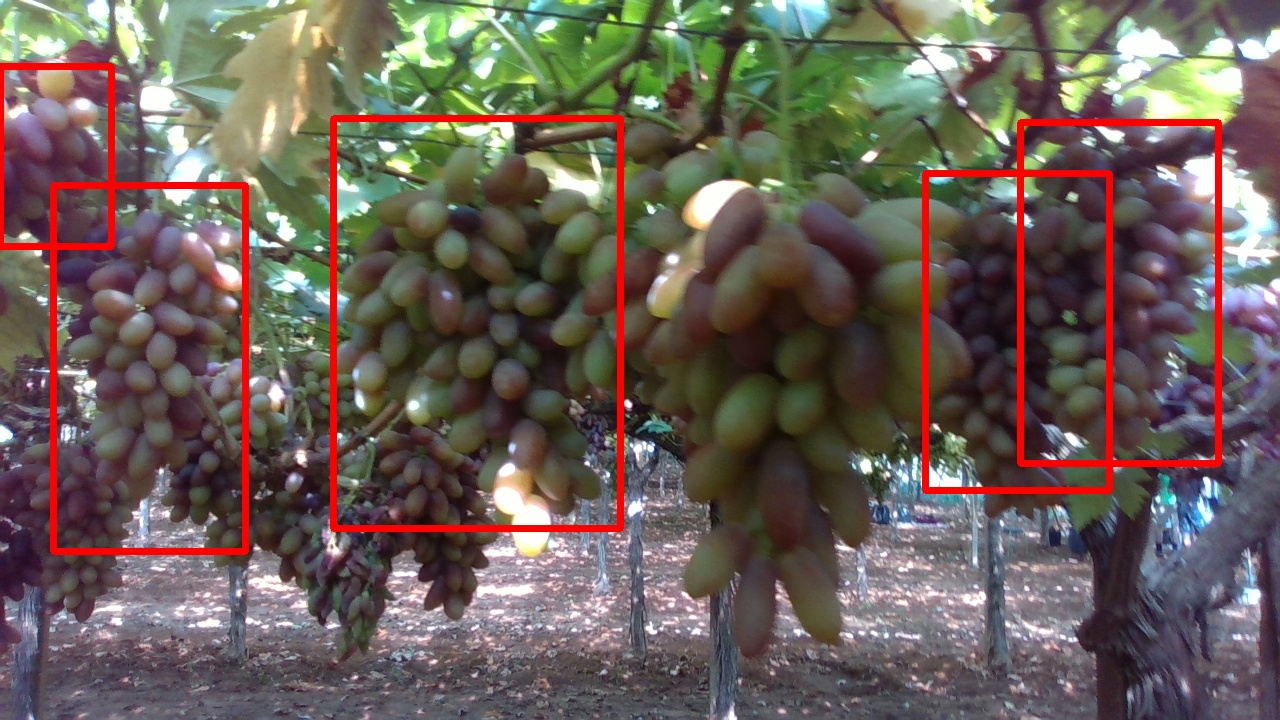}
    \end{subfigure}
    \begin{subfigure}[]{0.325\textwidth}
        \includegraphics[width=\columnwidth]{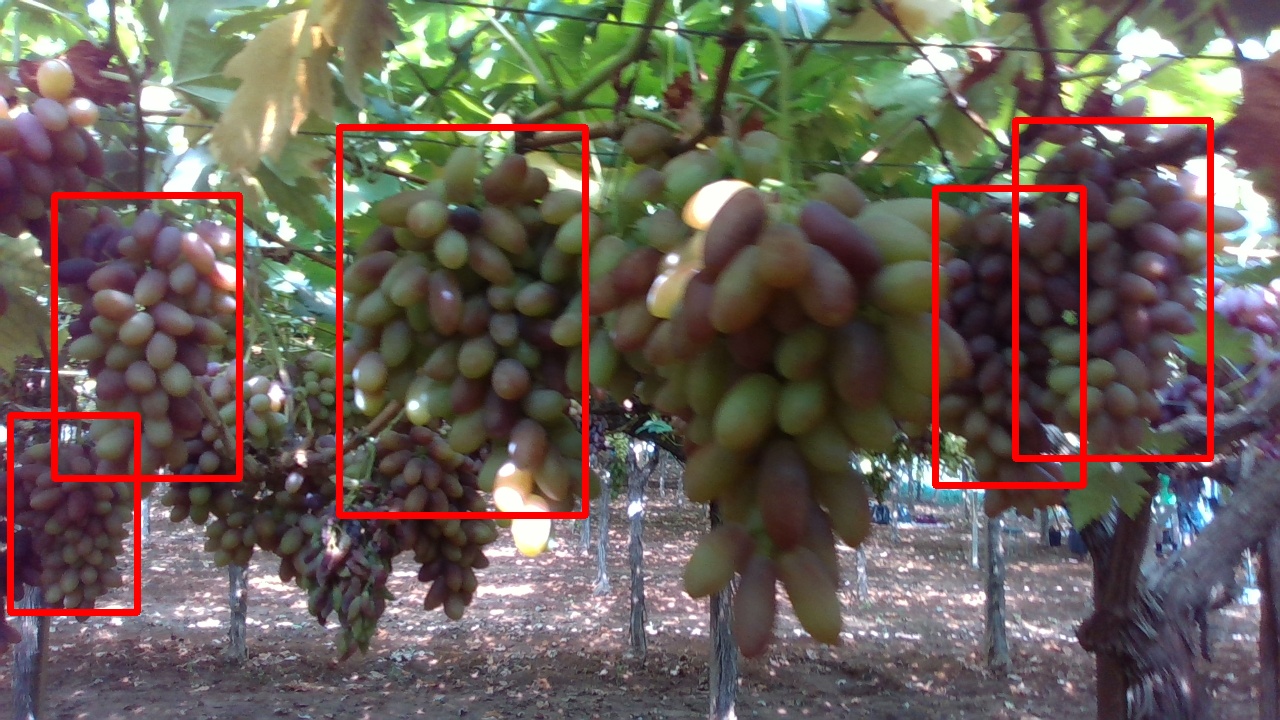}
    \end{subfigure}
    \begin{subfigure}[]{0.325\textwidth}
        \includegraphics[width=\columnwidth]{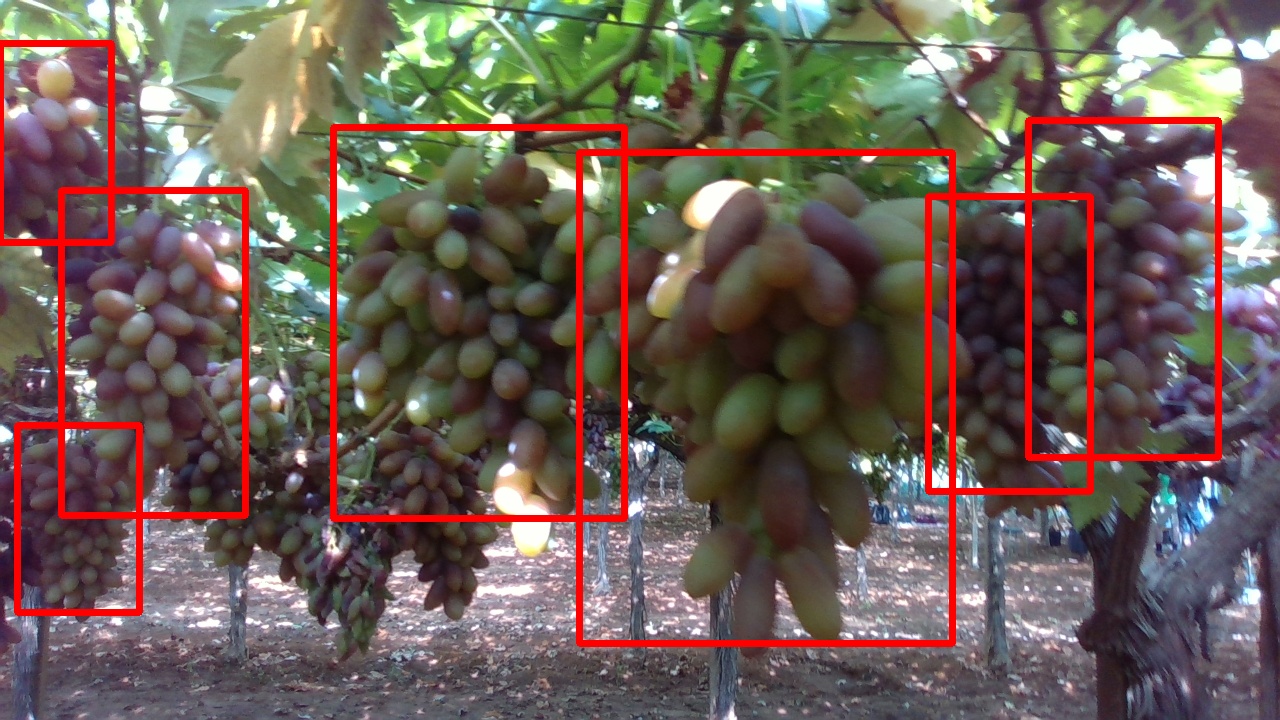}
    \end{subfigure}

    \vspace{5pt}
    \begin{subfigure}[]{0.325\textwidth}
        \includegraphics[width=\columnwidth]{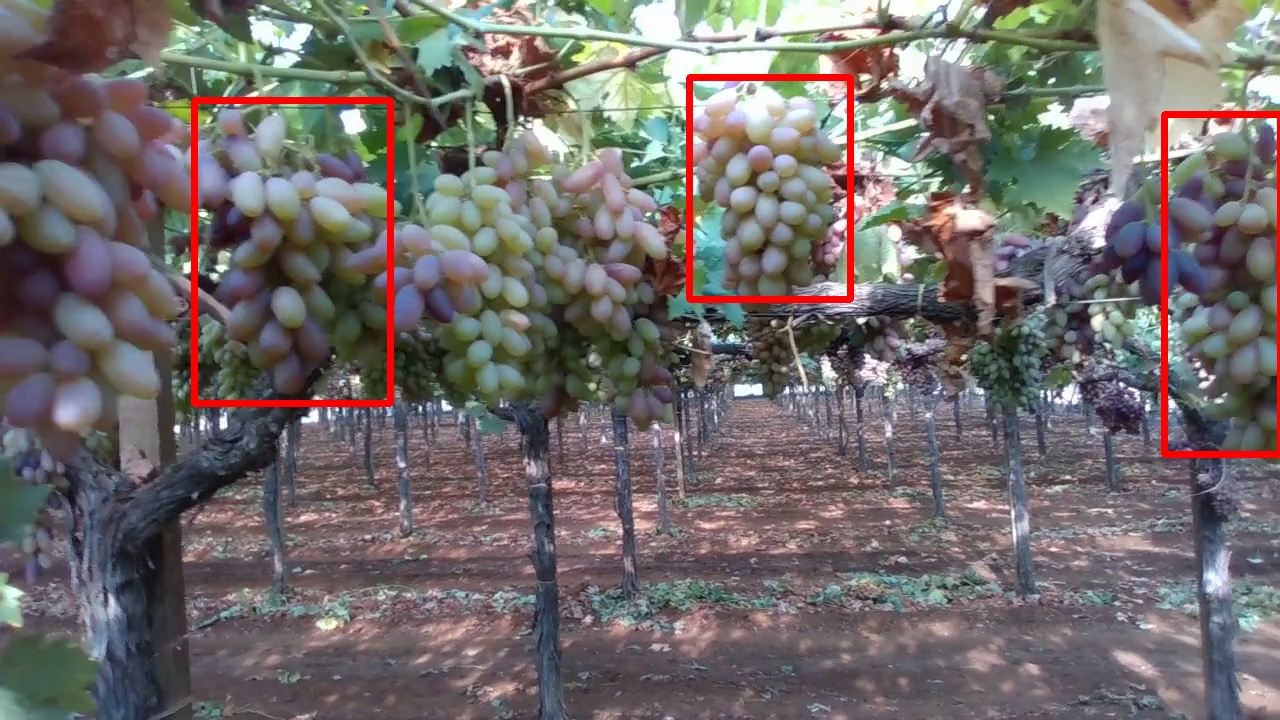}
        \caption{}
    \end{subfigure}
    \begin{subfigure}[]{0.325\textwidth}
        \includegraphics[width=\columnwidth]{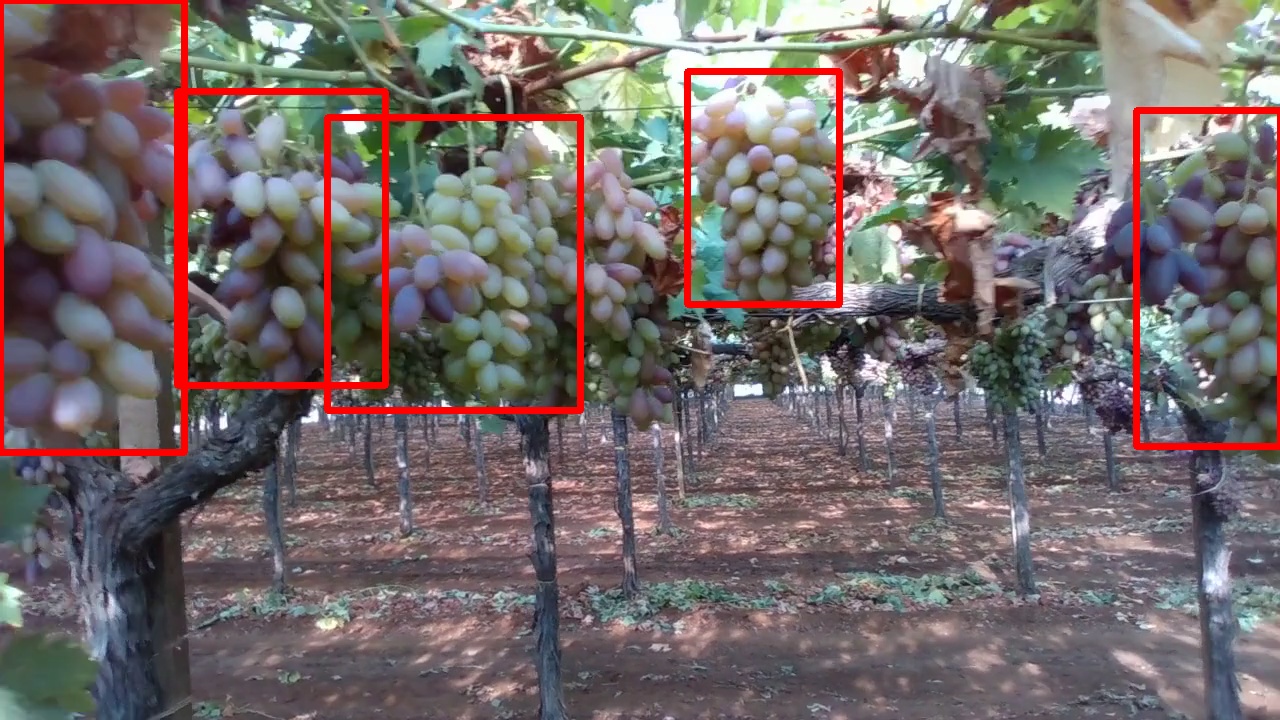}
        \caption{}
    \end{subfigure}
    \begin{subfigure}[]{0.325\textwidth}
        \includegraphics[width=\columnwidth]{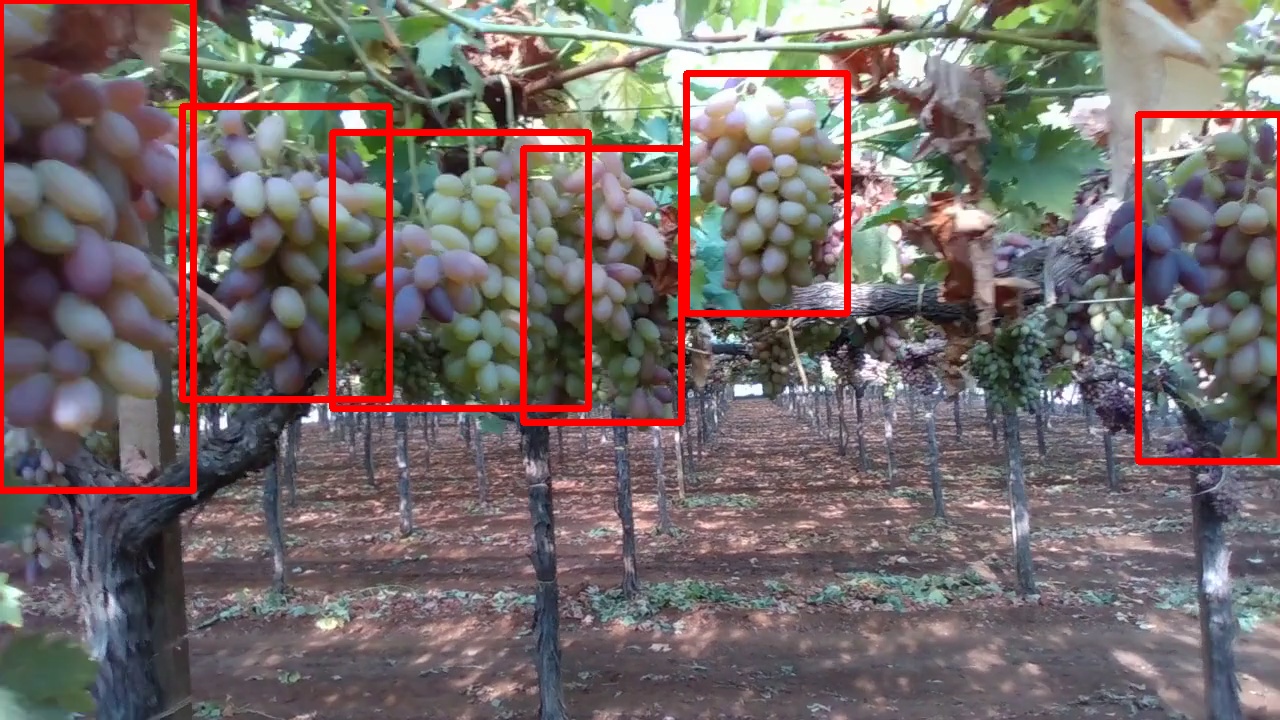}
        \caption{}
    \end{subfigure}
    \caption{\footnotesize Qualitative evaluation of the inference results of the YOLO nano models in frames extracted from different sequences used for the test using a confidence threshold of 0.25 and IoU of 0.3. On the first row is frame 39 extracted from the \textit{CloseUp1} sequence; on the second row, frame 3 from \textit{CloseUp2}; and on the third row, frame 5 from the \textit{Overview2} sequence. Each column corresponds to a different set used for training. The model trained using only the \textit{Pseudo} dataset (a) displays a very low recall with many false negatives due to its poor generalization capabilities. The model that uses the \textit{SyntheticPasted} set (b) shows an improvement compared to the baseline (\textit{Pseudo}). The best model is obtained by training using the \textit{SyntheticPasted + Pseudo} dataset (c), showing superior capabilities in cases of occlusions, intense illumination, and large clusters in the foreground.}
    \label{fig::comparison}
\end{figure*}

\begin{table*}[ht]
\centering
\caption{\footnotesize Comparison of different YOLOv8 versions trained on different data.}
\begin{adjustbox}{width=\textwidth}
\begin{tabular}{|c|c|ccccc|ccccc|}
\hline
 & Detector Model & \multicolumn{5}{c|}{YOLO \textit{nano}} & \multicolumn{5}{c|}{YOLO \textit{small}} \\ \hline
Sequence & Method & \multicolumn{1}{c|}{Precision↑} & \multicolumn{1}{c|}{Recall↑} & \multicolumn{1}{c|}{F1 Score↑} & \multicolumn{1}{c|}{mAP50↑} & mAP50-95↑ & \multicolumn{1}{c|}{Precision↑} & \multicolumn{1}{c|}{Recall↑} & \multicolumn{1}{c|}{F1 Score↑} & \multicolumn{1}{c|}{mAP50↑} & mAP50-95↑ \\ \hline
\multirow{5}{*}{CloseUp1} & Pseudo & \multicolumn{1}{c|}{\textbf{0,793}} & \multicolumn{1}{c|}{0,137} & \multicolumn{1}{c|}{0,233} & \multicolumn{1}{c|}{0,454} & 0,123 & \multicolumn{1}{c|}{\textbf{0,814}} & \multicolumn{1}{c|}{0,215} & \multicolumn{1}{c|}{0,34} & \multicolumn{1}{c|}{0,496} & 0,186 \\
 & Synthetic & \multicolumn{1}{c|}{0,431} & \multicolumn{1}{c|}{0,303} & \multicolumn{1}{c|}{0,356} & \multicolumn{1}{c|}{0,35} & 0,121 & \multicolumn{1}{c|}{0,557} & \multicolumn{1}{c|}{0,391} & \multicolumn{1}{c|}{0,459} & \multicolumn{1}{c|}{0,434} & 0,133 \\
 & Synthetic + Pseudo & \multicolumn{1}{c|}{0,696} & \multicolumn{1}{c|}{0,23} & \multicolumn{1}{c|}{0,346} & \multicolumn{1}{c|}{0,432} & 0,166 & \multicolumn{1}{c|}{0,72} & \multicolumn{1}{c|}{0,33} & \multicolumn{1}{c|}{0,452} & \multicolumn{1}{c|}{0,49} & 0,192 \\
 & SyntheticPasted & \multicolumn{1}{c|}{0,657} & \multicolumn{1}{c|}{\textbf{0,493}} & \multicolumn{1}{c|}{\textbf{0,563}} & \multicolumn{1}{c|}{\textbf{0,561}} & \textbf{0,237} & \multicolumn{1}{c|}{0,571} & \multicolumn{1}{c|}{\textbf{0,393}} & \multicolumn{1}{c|}{0,465} & \multicolumn{1}{c|}{0,474} & 0,192 \\
 & SyntheticPasted + Pseudo & \multicolumn{1}{c|}{0,781} & \multicolumn{1}{c|}{0,316} & \multicolumn{1}{c|}{0,45} & \multicolumn{1}{c|}{0,539} & 0,206 & \multicolumn{1}{c|}{0,777} & \multicolumn{1}{c|}{0,346} & \multicolumn{1}{c|}{\textbf{0,479}} & \multicolumn{1}{c|}{\textbf{0,527}} & \textbf{0,234} \\ \hline
\multirow{5}{*}{CloseUp2} & Pseudo & \multicolumn{1}{c|}{\textbf{0,989}} & \multicolumn{1}{c|}{0,297} & \multicolumn{1}{c|}{0,457} & \multicolumn{1}{c|}{0,628} & 0,392 & \multicolumn{1}{c|}{\textbf{0,965}} & \multicolumn{1}{c|}{0,322} & \multicolumn{1}{c|}{0,483} & \multicolumn{1}{c|}{0,647} & 0,406 \\
 & Synthetic & \multicolumn{1}{c|}{0,416} & \multicolumn{1}{c|}{0,388} & \multicolumn{1}{c|}{0,401} & \multicolumn{1}{c|}{0,38} & 0,165 & \multicolumn{1}{c|}{0,547} & \multicolumn{1}{c|}{0,524} & \multicolumn{1}{c|}{0,535} & \multicolumn{1}{c|}{0,531} & 0,257 \\
 & Synthetic + Pseudo & \multicolumn{1}{c|}{0,886} & \multicolumn{1}{c|}{0,368} & \multicolumn{1}{c|}{0,52} & \multicolumn{1}{c|}{0,626} & 0,375 & \multicolumn{1}{c|}{0,863} & \multicolumn{1}{c|}{0,446} & \multicolumn{1}{c|}{0,588} & \multicolumn{1}{c|}{\textbf{0,672}} & \textbf{0,4} \\
 & SyntheticPasted & \multicolumn{1}{c|}{0,609} & \multicolumn{1}{c|}{0,454} & \multicolumn{1}{c|}{0,52} & \multicolumn{1}{c|}{0,548} & 0,317 & \multicolumn{1}{c|}{0,615} & \multicolumn{1}{c|}{\textbf{0,557}} & \multicolumn{1}{c|}{0,584} & \multicolumn{1}{c|}{0,599} & 0,324 \\
 & SyntheticPasted + Pseudo & \multicolumn{1}{c|}{0,908} & \multicolumn{1}{c|}{\textbf{0,462}} & \multicolumn{1}{c|}{\textbf{0,612}} & \multicolumn{1}{c|}{\textbf{0,701}} & \textbf{0,433} & \multicolumn{1}{c|}{0,838} & \multicolumn{1}{c|}{0,502} & \multicolumn{1}{c|}{\textbf{0,628}} & \multicolumn{1}{c|}{0,665} & 0,42 \\ \hline
\multirow{5}{*}{Overview1} & Pseudo & \multicolumn{1}{c|}{\textbf{0,95}} & \multicolumn{1}{c|}{0,544} & \multicolumn{1}{c|}{\textbf{0,692}} & \multicolumn{1}{c|}{\textbf{0,741}} & \textbf{0,348} & \multicolumn{1}{c|}{\textbf{0,911}} & \multicolumn{1}{c|}{\textbf{0,582}} & \multicolumn{1}{c|}{\textbf{0,71}} & \multicolumn{1}{c|}{0,746} & \textbf{0,377} \\
 & Synthetic & \multicolumn{1}{c|}{0,0965} & \multicolumn{1}{c|}{0,147} & \multicolumn{1}{c|}{0,116} & \multicolumn{1}{c|}{0,0553} & 0,0165 & \multicolumn{1}{c|}{0,463} & \multicolumn{1}{c|}{0,308} & \multicolumn{1}{c|}{0,37} & \multicolumn{1}{c|}{0,285} & 0,0844 \\
 & Synthetic + Pseudo & \multicolumn{1}{c|}{0,883} & \multicolumn{1}{c|}{0,483} & \multicolumn{1}{c|}{0,624} & \multicolumn{1}{c|}{0,649} & 0,287 & \multicolumn{1}{c|}{0,901} & \multicolumn{1}{c|}{0,585} & \multicolumn{1}{c|}{0,709} & \multicolumn{1}{c|}{\textbf{0,752}} & 0,367 \\
 & SyntheticPasted & \multicolumn{1}{c|}{0,44} & \multicolumn{1}{c|}{0,425} & \multicolumn{1}{c|}{0,432} & \multicolumn{1}{c|}{0,369} & 0,146 & \multicolumn{1}{c|}{0,486} & \multicolumn{1}{c|}{0,358} & \multicolumn{1}{c|}{0,416} & \multicolumn{1}{c|}{0,366} & 0,157 \\
 & SyntheticPasted + Pseudo & \multicolumn{1}{c|}{0,879} & \multicolumn{1}{c|}{\textbf{0,557}} & \multicolumn{1}{c|}{0,681} & \multicolumn{1}{c|}{0,706} & 0,309 & \multicolumn{1}{c|}{0,845} & \multicolumn{1}{c|}{0,554} & \multicolumn{1}{c|}{0,669} & \multicolumn{1}{c|}{0,667} & 0,35 \\ \hline
\multirow{5}{*}{Overview2} & Pseudo & \multicolumn{1}{c|}{\textbf{0,877}} & \multicolumn{1}{c|}{0,297} & \multicolumn{1}{c|}{0,444} & \multicolumn{1}{c|}{0,593} & 0,301 & \multicolumn{1}{c|}{\textbf{0,854}} & \multicolumn{1}{c|}{0,43} & \multicolumn{1}{c|}{0,572} & \multicolumn{1}{c|}{\textbf{0,645}} & \textbf{0,312} \\
 & Synthetic & \multicolumn{1}{c|}{0,328} & \multicolumn{1}{c|}{0,266} & \multicolumn{1}{c|}{0,294} & \multicolumn{1}{c|}{0,25} & 0,0818 & \multicolumn{1}{c|}{0,308} & \multicolumn{1}{c|}{0,359} & \multicolumn{1}{c|}{0,331} & \multicolumn{1}{c|}{0,312} & 0,13 \\
 & Synthetic + Pseudo & \multicolumn{1}{c|}{0,768} & \multicolumn{1}{c|}{0,449} & \multicolumn{1}{c|}{0,566} & \multicolumn{1}{c|}{0,611} & 0,287 & \multicolumn{1}{c|}{0,776} & \multicolumn{1}{c|}{0,423} & \multicolumn{1}{c|}{0,547} & \multicolumn{1}{c|}{0,616} & 0,284 \\
 & SyntheticPasted & \multicolumn{1}{c|}{0,522} & \multicolumn{1}{c|}{0,42} & \multicolumn{1}{c|}{0,465} & \multicolumn{1}{c|}{0,481} & 0,244 & \multicolumn{1}{c|}{0,457} & \multicolumn{1}{c|}{0,419} & \multicolumn{1}{c|}{0,437} & \multicolumn{1}{c|}{0,377} & 0,181 \\
 & SyntheticPasted + Pseudo & \multicolumn{1}{c|}{0,758} & \multicolumn{1}{c|}{\textbf{0,546}} & \multicolumn{1}{c|}{\textbf{0,635}} & \multicolumn{1}{c|}{\textbf{0,667}} & \textbf{0,336} & \multicolumn{1}{c|}{0,672} & \multicolumn{1}{c|}{\textbf{0,544}} & \multicolumn{1}{c|}{\textbf{0,601}} & \multicolumn{1}{c|}{0,614} & 0,298 \\ \hline
\end{tabular}
\end{adjustbox}
\label{tab::performance}
\vspace{-5.0mm}
\end{table*}

\subsection{Real World to Simulator Cut-and-Paste}
\label{sec::paste}
The last step of the proposed method consists of pasting the extracted real instances into the simulation environment. Rather than employing random pasting, as seen in Cut-and-Paste augmentation \cite{dwibedi2017cut}, we adopt a more structured approach. Using the synthetic position of the bunches allows for a more realistic positioning of the grapes compared to random pasting. We randomly sample a real mask from $B_{real}$, resize and rotate it, and then overlay it onto a synthetic bunch instance. Since the simulator has the same texture for each table grape bunch, covering it with a natural texture is needed to generate photo-realistic images and maintain variability in the input distribution.

To compensate for the orientation difference between the real bunch mask $M_{real}$ and the simulated bunch mask $M_{sim}$, we perform Principal Component Analysis ($PCA$), extracting the first two principal components ($P_{real}$ and $P_{sim}$) corresponding to the longest and shortest axes of the bunch Eq. (\ref{eq::rotation}). 

\begin{equation}
\begin{split}
\label{eq::rotation}
P_{real} &= PCA(M_{real}) \\
P_{sim} &= PCA(M_{sim})\\
\end{split}
\end{equation}

Then, we align the real mask with the simulated one using the minimum angle of rotation $\theta_{rot}$ between the two principal axes, considering both senses of rotation using Eq. (\ref{eq::rot}).

\begin{equation}
\begin{split}
\label{eq::rot}
\theta_{real} =& atan_2(P_{real})\\
\theta1_{sim} =& atan_2(P_{sim})\\
\theta2_{sim} =& atan_2(-P_{sim})\\
\theta_{rot} = min(|\theta_{real} - \theta&1_{sim}|, |\theta_{real} - \theta2_{sim}|)\\
\end{split}
\end{equation}

To compensate for the scale difference, we take the smallest bounding box that fits all points of the segmentation masks and calculate the height and width scale factors as follows:

\begin{equation}
\begin{split}
\label{eq::scale_factors}
f_w &= sim_w / real_w \\
f_h &= sim_h / real_h\\
\end{split}
\xrightarrow{\text{}} f = max(f_w, f_h)
\end{equation}

where $sim_w$, $sim_h$, $real_w$ and $real_h$ are the mask dimensions and $f_w, f_h$ are the width and height scaling factors. 
Finally, we resize the image using the maximum scaling factor to avoid texture deformation and to ensure that the real texture covers most of the synthetic bunch in the virtual environment. The last step before blending is clipping the real mask by taking only the parts common to the synthetic one. By doing so, it is possible to preserve occlusions due to other grape bunches, leaves, and branches, giving more realism to the scene.

\section{Experiments}
\label{sec::experiments}
The experiments described in the following Section show how combining real and synthetic data using a cut-and-paste technique can improve the performance of a detector trained only using automatically generated pseudo-labels \cite{ciarfuglia2023weakly, ciarfuglia2022pseudo}.

\subsection{Training Details}
\label{sec::training_details}
The state-of-the-art object detection algorithm chosen for the experiments is YOLOv8, provided by Ultralytics \cite{yolov8_ultralytics}. We conduct all experiments on a workstation equipped with a NVIDIA GeForce RTX 3090 GPU with 24 GB of VRAM, an AMD Ryzen 9 5950 16-core processor, and 64 GB of RAM. All the training runs consist of 100 epochs with a batch size of 8, and we employ early stopping with a patience parameter of 50 epochs. We use the one-cycle learning rate (\textit{lr}) policy \cite{smith2019super}, with initial \textit{lr} = \num{1e-4} and final \textit{lr} = \num{1e-6}. The optimizer is SGD, with a momentum of 0.937 and a weight decay of \num{5e-4}. The training time is highly variable and depends on the number of images used and the version of YOLO. In particular, five different versions can be used, with parameter counts ranging from 3.2 million to 11.2 million for the \textit{nano} and \textit{small} variants, and reaching 68.2 million for the \textit{extra large} version. 
This work uses only the \textit{nano} and \textit{small} versions since those are the ones with the least parameters and can be deployed with real-time usage on a limited hardware robotic platform. All the models are pretrained on the COCO dataset \cite{lin2014microsoft} and then finetuned on our data. We trained the baseline model only on the pseudo labels dataset. In contrast, we trained the proposed models on the synthetic and cut-and-paste set generated using the method described in \ref{sec::paste}. We do not apply any augmentation technique to the images in the dataset to ensure that the results are only due to the introduction of synthetic data. 

\subsection{Evaluation Metrics}
\label{sec::evaluation_metrics}
We evaluate our experiments using a number of accepted metrics for the object detection problems \cite{9145130}\cite{lin2014microsoft}, namely Precision, Recall, F1-Score, Intersection over Union (IoU), and mean Average Precision (mAP). Precision is the proportion of true positives among all positive predictions, assessing the model's capability to detect objects correctly. On the other hand, Recall is the ratio of true positive instances to the total number of positive instances in the dataset, indicating the model's ability to identify all instances of objects in the images. The F1 Score is the harmonic mean of Precision and Recall (Eq. \ref{eq::f1score}) and provides a balanced assessment of a model's performance considering both false positives and false negatives.

\begin{equation}
\label{eq::f1score}
F1 = 2 \times \frac{P \times R}{P + R}
\end{equation}

The Average Precision (AP) computes the area under the precision-recall curve, which is usually computed for each class and then averaged to obtain the mean average precision. In our case, AP coincides with the mAP because we consider only one class (grape). The mAP50 is defined using an IoU threshold of 0.5, which indicates the model performance considering only relatively easy detections. In contrast, mAP50-95 is computed at ten IoU thresholds from $0.5$ to $0.95$ with a step size of $0.05$.

\subsection{Results and discussion}
\label{sec::results}
We present the performance of the detectors trained using data generated with the method proposed in this paper in Table \ref{tab::performance}. We compare the nano and small versions of the YOLOv8 detector trained on five different training splits:
\begin{itemize}
\item \textbf{Pseudo}: pseudo-labels dataset presented in \cite{ciarfuglia2023weakly} which consists of 2400 images with labels automatically generated from video sequences taken in 2021 with a RealSense d435i camera.
\item \textbf{Synthetic}: dataset consisting of 1889 images captured using the simulator, with the ground truth segmentation masks generated by the simulator. We captured images with different illumination conditions as described in Sec. \ref{sec::data}: 209 with low light, 707 with medium light, 559 with high light, and 416 under very strong backlighting. 
\item \textbf{Synthetic + Pseudo}: the union of the \textit{Synthetic} and \textit{Pseudo} dataset, resulting in 4289 images.
\item \textbf{SyntheticPasted}: dataset containing the same images of the \textit{Synthetic} dataset but with the real instances pasted using the proposed method described in Sec. \ref{sec::methods}.
\item \textbf{SyntheticPasted + Pseudo}: the combination of the \textit{SyntheticPasted} and \textit{Pseudo} datasets, therefore contains 4289 images.
\end{itemize}

We present a visual representation of the difference in the performance among the detectors trained using the dataset proposed in this work in Fig. \ref{fig::comparison}. The data used for testing have been presented in \cite{saraceni2023agrisort} as a benchmark for Multiple Object Tracking of table grapes, but we re-adapted it for Object Detection. The \textit{CloseUp} sequences consist of 300 frames each, captured in 2023 with the RealSense d435i camera. The \textit{Overview} sequences consist of 100 frames each, captured in 2021, with a more distant perspective than the \textit{CloseUps} and are more similar to the images in the \textit{Pseudo} dataset. The covariate shift between the \textit{Pseudo} training set and those sequences is introduced by the different years of capturing, illumination conditions, occlusions, and distance of the camera from the crops. 

The baseline consists of the YOLOv8 detectors trained using only the \textit{Pseudo} set. From the results, it is clear that those models obtain the highest Precision for both small and nano versions of the detector, meaning that its bounding box prediction is tight around the instances. However, the Recall and mAP results are poor in most sequences. The results show that introducing purely synthetic data (\textit{Synthetic}) improves Recall in some sequences, allowing the model to detect more instances and showing the potential of using a simulator to train a detector to improve generalizability. The model trained using the \textit{SyntheticPasted} set performs better in all metrics than purely synthetic data, underlining that blending real instances into a synthetic scenario using the methodology proposed brings positive effects compared to simply using synthetic data. However, using only synthetic data (with or without Pasting) results in a consistent loss in Precision and mAP compared to the baseline. The models produce many false positives because the images captured by the simulator are far from photo-realistic. In particular, the simulation environment is characterized by dense foliage, which causes many occlusions by leaves and branches, leading to poor Precision. By combining the \textit{Synthetic} and \textit{Pseudo} datasets to train the detector, we obtain the best of both worlds, improving both Recall and mAP values and surpassing the baseline. In particular, the effect obtained mitigates the problem of the high number of false positive detections in the \textit{Synthetic} and \textit{SyntheticPasted} cases. The detector can better distinguish single clusters in cluttered cases and does not confuse the background for grapes. The \textit{SyntheticPasted + Pseudo} model (Fig. \ref{fig::comparison}.c) obtains better results compared to the \textit{Synthetic + Pseudo} (Fig. \ref{fig::comparison}.b), mainly because it can identify clusters closer to the camera and disambiguate multiple clusters better when they occlude each other.

\section{Conclusions}
\label{sec::conclusions}
In this paper, we present a pipeline to improve the performance of a YOLOv8 detector for agricultural purposes by generating photo-realistic data from a simulator. We automatically extract grape segmentation masks from real photos using a combination of a given base detector (Yolov5) and SAM. Then, we blend the real instances onto synthetic images from an agricultural simulated environment using a cut-and-paste approach. We employ geometrical consistency considerations by rotating, rescaling, and translating the real instances to align them to the simulated ones to improve the quality of the synthetic images. Our experiments demonstrated that the detector trained using data generated with the method described in this work shows considerable improvement over the baseline. The model better disambiguates cluttered instances and exhibits more robustness towards occlusions, illumination changes, and motion blur. 

\bibliographystyle{ieeetr}
\bibliography{mybibliography}

\begin{thebibliography}{10}

\bibitem{cheng2023recent}
C.~Cheng, J.~Fu, H.~Su, and L.~Ren, ``Recent advancements in agriculture
  robots: Benefits and challenges,'' {\em Machines}, vol.~11, no.~1, p.~48,
  2023.

\bibitem{droukas2023survey}
L.~Droukas, Z.~Doulgeri, N.~L. Tsakiridis, D.~Triantafyllou, I.~Kleitsiotis,
  I.~Mariolis, D.~Giakoumis, D.~Tzovaras, D.~Kateris, and D.~Bochtis, ``A
  survey of robotic harvesting systems and enabling technologies,'' {\em
  Journal of Intelligent \& Robotic Systems}, vol.~107, no.~2, p.~21, 2023.

\bibitem{kurtser2020in-field}
P.~Kurtser, O.~Ringdahl, N.~Rotstein, R.~Berenstein, and Y.~Edan, ``In-field
  grape cluster size assessment for vine yield estimation using a mobile robot
  and a consumer level rgb-d camera,'' {\em IEEE Robotics and Automation
  Letters}, vol.~5, no.~2, pp.~2031--2038, 2020.

\bibitem{lacotte2022pesticide-free}
V.~Lacotte, T.~NGuyen, J.~D. Sempere, V.~Novales, V.~Dufour, R.~Moreau, M.~T.
  Pham, K.~Rabenorosoa, S.~Peignier, F.~G. Feugier, R.~Gaetani, T.~Grenier,
  B.~Masenelli, P.~da~Silva, A.~Heddi, and A.~Lelevé, ``Pesticide-free robotic
  control of aphids as crop pests,'' {\em AgriEngineering}, vol.~4, no.~4,
  pp.~903--921, 2022.

\bibitem{pretto2020building}
A.~Pretto, S.~Aravecchia, W.~Burgard, N.~Chebrolu, C.~Dornhege, T.~Falck,
  F.~Fleckenstein, A.~Fontenla, M.~Imperoli, R.~Khanna, F.~Liebisch, P.~Lottes,
  A.~Milioto, D.~Nardi, S.~Nardi, J.~Pfeifer, M.~Popović, C.~Potena,
  C.~Pradalier, E.~Rothacker-Feder, I.~Sa, A.~Schaefer, R.~Siegwart,
  C.~Stachniss, A.~Walter, W.~Winterhalter, X.~Wu, and J.~Nieto, ``Building an
  aerial–ground robotics system for precision farming: An adaptable
  solution,'' {\em IEEE Robotics \& Automation Magazine}, vol.~28, no.~3,
  pp.~29--49, 2021.

\bibitem{lippi2023autonomous}
M.~Lippi, M.~Santilli, R.~F. Carpio, J.~Maiolini, E.~Garone, V.~Cristofori, and
  A.~Gasparri, ``An autonomous spraying robot architecture for sucker
  management in large-scale hazelnut orchards,'' {\em Journal of Field
  Robotics}.

\bibitem{stavridis2121pick-and-place}
S.~Stavridis, P.~Falco, and Z.~Doulgeri, ``Pick-and-place in dynamic
  environments with a mobile dual-arm robot equipped with distributed distance
  sensors,'' in {\em 2020 IEEE-RAS 20th International Conference on Humanoid
  Robots (Humanoids)}, pp.~76--82, 2021.

\bibitem{ginart2022mldemon}
T.~Ginart, M.~Jinye~Zhang, and J.~Zou, ``Mldemon:deployment monitoring for
  machine learning systems,'' in {\em Proceedings of The 25th International
  Conference on Artificial Intelligence and Statistics} (G.~Camps-Valls,
  F.~J.~R. Ruiz, and I.~Valera, eds.), vol.~151 of {\em Proceedings of Machine
  Learning Research}, pp.~3962--3997, PMLR, 28--30 Mar 2022.

\bibitem{park2021reliable}
C.~Park, A.~Awadalla, T.~Kohno, and S.~Patel, ``Reliable and trustworthy
  machine learning for health using dataset shift detection,'' in {\em Advances
  in Neural Information Processing Systems} (M.~Ranzato, A.~Beygelzimer,
  Y.~Dauphin, P.~Liang, and J.~W. Vaughan, eds.), vol.~34, pp.~3043--3056,
  Curran Associates, Inc., 2021.

\bibitem{canopies}
``Eu canopies project: A collaborative paradigm for human workers and
  multi-robot teams in precision agriculture systems.''
  \url{https://canopies.inf.uniroma3.it/}, 2021.

\bibitem{roscher2023data}
R.~Roscher, L.~Roth, C.~Stachniss, and A.~Walter, ``Data-centric digital
  agriculture: A perspective,'' {\em arXiv preprint arXiv:2312.03437}, 2023.

\bibitem{kamilaris2017review}
A.~Kamilaris, A.~Kartakoullis, and F.~X. Prenafeta-Bold{\'u}, ``A review on the
  practice of big data analysis in agriculture,'' {\em Computers and
  electronics in agriculture}, vol.~143, pp.~23--37, 2017.

\bibitem{li2023label}
J.~Li, D.~Chen, X.~Qi, Z.~Li, Y.~Huang, D.~Morris, and X.~Tan,
  ``Label-efficient learning in agriculture: A comprehensive review,'' {\em
  Computers and Electronics in Agriculture}, vol.~215, p.~108412, 2023.

\bibitem{goodfellow2020generative}
I.~Goodfellow, J.~Pouget-Abadie, M.~Mirza, B.~Xu, D.~Warde-Farley, S.~Ozair,
  A.~Courville, and Y.~Bengio, ``Generative adversarial networks,'' {\em
  Commun. ACM}, vol.~63, p.~139–144, oct 2020.

\bibitem{lu2022generative}
Y.~Lu, D.~Chen, E.~Olaniyi, and Y.~Huang, ``Generative adversarial networks
  (gans) for image augmentation in agriculture: A systematic review,'' {\em
  Computers and Electronics in Agriculture}, vol.~200, p.~107208, 2022.

\bibitem{madsen2019generating}
S.~L. Madsen, M.~Dyrmann, R.~N. Jørgensen, and H.~Karstoft, ``Generating
  artificial images of plant seedlings using generative adversarial networks,''
  {\em Biosystems Engineering}, vol.~187, pp.~147--159, 2019.

\bibitem{valerio2017arigan}
M.~Valerio~Giuffrida, H.~Scharr, and S.~A. Tsaftaris, ``Arigan: Synthetic
  arabidopsis plants using generative adversarial network,'' in {\em
  Proceedings of the IEEE international conference on computer vision
  workshops}, pp.~2064--2071, 2017.

\bibitem{hammouch2022ganset}
H.~Hammouch, S.~Mohapatra, M.~El-Yacoubi, H.~Qin, H.~Berbia, P.~Mäder, and
  M.~Chikhaoui, ``Ganset - generating annnotated datasets using generative
  adversarial networks,'' in {\em 2022 International Conference on
  Cyber-Physical Social Intelligence (ICCSI)}, pp.~615--620, 2022.

\bibitem{karam2022gan-based}
C.~Karam, M.~Awad, Y.~Abou~Jawdah, N.~Ezzeddine, and A.~Fardoun, ``Gan-based
  semi-automated augmentation online tool for agricultural pest detection: A
  case study on whiteflies,'' {\em Frontiers in Plant Science}, vol.~13, 2022.

\bibitem{paulin2023review}
G.~Paulin and M.~Ivasic-Kos, ``Review and analysis of synthetic dataset
  generation methods and techniques for application in computer vision,'' {\em
  Artificial Intelligence Review}, vol.~56, no.~9, pp.~9221--9265, 2023.

\bibitem{barth2018data}
R.~Barth, J.~IJsselmuiden, J.~Hemming, and E.~V. Henten, ``Data synthesis
  methods for semantic segmentation in agriculture: A capsicum annuum
  dataset,'' {\em Computers and Electronics in Agriculture}, vol.~144,
  pp.~284--296, 2018.

\bibitem{hartley2021domain}
Z.~K.~J. Hartley and A.~P. French, ``Domain adaptation of synthetic images for
  wheat head detection,'' {\em Plants}, vol.~10, no.~12, 2021.

\bibitem{barth2018improved}
R.~Barth, J.~Hemming, and E.~J. van Henten, ``Improved part segmentation
  performance by optimising realism of synthetic images using cycle generative
  adversarial networks,'' {\em arXiv preprint arXiv:1803.06301}, 2018.

\bibitem{zhu2017unpaired}
J.-Y. Zhu, T.~Park, P.~Isola, and A.~A. Efros, ``Unpaired image-to-image
  translation using cycle-consistent adversarial networks,'' in {\em Computer
  Vision (ICCV), 2017 IEEE International Conference on}, 2017.

\bibitem{canopiesD2.4}
{PaleBlue AS}, {\em VR Farming Environment Specification}.
\newblock The CANOPIES Project, 2014.

\bibitem{ciarfuglia2023weakly}
T.~A. Ciarfuglia, I.~M. Motoi, L.~Saraceni, M.~Fawakherji, A.~Sanfeliu, and
  D.~Nardi, ``Weakly and semi-supervised detection, segmentation and tracking
  of table grapes with limited and noisy data,'' {\em Computers and Electronics
  in Agriculture}, vol.~205, p.~107624, 2023.

\bibitem{ciarfuglia2022pseudo}
T.~A. Ciarfuglia, I.~M. Motoi, L.~Saraceni, and D.~Nardi, ``Pseudo-label
  generation for agricultural robotics applications,'' in {\em Proceedings of
  the IEEE/CVF Conference on Computer Vision and Pattern Recognition},
  pp.~1686--1694, 2022.

\bibitem{kirillov2023segment}
A.~Kirillov, E.~Mintun, N.~Ravi, H.~Mao, C.~Rolland, L.~Gustafson, T.~Xiao,
  S.~Whitehead, A.~C. Berg, W.-Y. Lo, {\em et~al.}, ``Segment anything,'' {\em
  arXiv preprint arXiv:2304.02643}, 2023.

\bibitem{dwibedi2017cut}
D.~Dwibedi, I.~Misra, and M.~Hebert, ``Cut, paste and learn: Surprisingly easy
  synthesis for instance detection,'' in {\em Proceedings of the IEEE
  international conference on computer vision}, pp.~1301--1310, 2017.

\bibitem{yolov8_ultralytics}
G.~Jocher, A.~Chaurasia, and J.~Qiu, ``Ultralytics yolov8,'' 2023.

\bibitem{smith2019super}
L.~N. Smith and N.~Topin, ``Super-convergence: Very fast training of neural
  networks using large learning rates,'' in {\em Artificial intelligence and
  machine learning for multi-domain operations applications}, vol.~11006,
  pp.~369--386, SPIE, 2019.

\bibitem{lin2014microsoft}
T.-Y. Lin, M.~Maire, S.~Belongie, J.~Hays, P.~Perona, D.~Ramanan,
  P.~Doll{\'a}r, and C.~L. Zitnick, ``Microsoft coco: Common objects in
  context,'' in {\em Computer Vision--ECCV 2014: 13th European Conference,
  Zurich, Switzerland, September 6-12, 2014, Proceedings, Part V 13},
  pp.~740--755, Springer, 2014.

\bibitem{9145130}
R.~Padilla, S.~L. Netto, and E.~A.~B. da~Silva, ``A survey on performance
  metrics for object-detection algorithms,'' in {\em 2020 International
  Conference on Systems, Signals and Image Processing (IWSSIP)}, pp.~237--242,
  2020.

\bibitem{saraceni2023agrisort}
L.~Saraceni, I.~M. Motoi, D.~Nardi, and T.~A. Ciarfuglia, ``Agrisort: A simple
  online real-time tracking-by-detection framework for robotics in precision
  agriculture,'' {\em arXiv preprint arXiv:2309.13393}, 2023.

\end{thebibliography}

\end{document}